\documentclass{article}

\usepackage{arxiv}

\usepackage[utf8]{inputenc} 
\usepackage[T1]{fontenc}    
\usepackage{hyperref}       
\usepackage{url}            
\usepackage{booktabs}       
\usepackage{amsfonts}       
\usepackage{nicefrac}       
\usepackage{microtype}      
\usepackage{lipsum}		
\usepackage{graphicx}
\usepackage{natbib}
\usepackage{doi}
\usepackage{graphicx}

\title{IDENTIFYING NEGATIVITY FACTORS FROM SOCIAL MEDIA TEXT CORPUS USING SENTIMENT ANALYSIS METHOD}

\author{\author{Mohammad Aimal$^{1}$  \and
        Maheen Bakhtyar$^{1}$ \and
        Junaid Baber$^{1}$ \and
        Sadia Lakho$^{1}$ \and
        Umar Mohammad$^{1}$ \and
        Warda Ahmed $^{2}$ \and 
        Jahanvash Karim$^{3}$
}
\thanks{This is the pre-print version of the paper accepted in SOFA 2020 Conference.} \\
\And
{\includegraphics[scale=0.06]{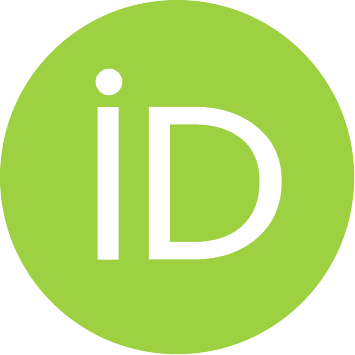}\hspace{1mm}Mohammad Aimal} \\
	Department of CS and IT\\
	University of Balochistan\\	
	\texttt{malikaimal090@gmail.com} \\
	\And
	{\includegraphics[scale=0.06]{orcid.pdf}\hspace{1mm} Maheen Bakhtyar} \\
	Department of CS and IT\\
	University of Balochistan\\
	\texttt{maheen.bakhtyar@um.uob.edu.pk} \\
\And
{\includegraphics[scale=0.06]{orcid.pdf}\hspace{1mm}Junaid Baber} \\
	Department of CS and IT\\
	University of Balochistan\\	
	\texttt{junaidbaber@ieee.org} \\
	\And
	{\includegraphics[scale=0.06]{orcid.pdf}\hspace{1mm} Sadia Lakho} \\
	Department of CS and IT\\
	SBK Women's University Balochistan\\
	\texttt{sadia.lakho@gmail.com} \\
\And
	{\includegraphics[scale=0.06]{orcid.pdf}\hspace{1mm} Wardah Ahmed} \\
	Department of CS and IT\\
	SBK Women's University Balochistan\\
	\texttt{wardahahmed12@gmail.com} \\
\And
{\includegraphics[scale=0.06]{orcid.pdf}\hspace{1mm}Jahanvash Karim} \\
	Institute of Management Sciences\\
	University of Balochistan\\	
	\texttt{ j\_vash@hotmail.com} \\
}

\renewcommand{\shorttitle}{\textit{arXiv} Template}

\hypersetup{
pdftitle={A template for the arxiv style},
pdfsubject={q-bio.NC, q-bio.QM},
pdfauthor={David S.~Hippocampus, Elias D.~Striatum},
pdfkeywords={First keyword, Second keyword, More},
}

\begin{document}
\maketitle

\begin{abstract}
	
Automatic sentiment analysis play vital role in decision making. Many organizations spend a lot of budget to understand their customer satisfaction by manually going over their feedback/comments or tweets. Automatic sentiment analysis can give overall picture of the comments received against any event, product, or activity. 
Usually, the comments/tweets are classified into two main classes that are negative or positive. However, the negative comments are too abstract to understand the basic reason or the context. organizations are interested to identify the exact reason for the negativity.  
In this research study, we hierarchically goes down into negative comments, and link them with more classes. Tweets are extracted from social media sites such as Twitter and Facebook. If the sentiment analysis classifies any tweet into negative class, then we further try to associates that negative comments with more possible negative classes. Based on expert opinions, the negative comments/tweets are further classified into 8 classes. Different machine learning algorithms are evaluated and their accuracy are reported. 
\end{abstract}


\section{Introduction}
In this technology area, where we are connected every second to the globe using internet/ social media, and with the thumb scrolling we can share and know the experiences. The data on the internet/ social media is increasing at an exponential rate. Now most of the time whatever we read or write is unfiltered and moderated which creates or may create drastic circumstances for individuals, groups of people, societies, even for countries. For example, discussion on Religion mostly results in violence, particularly in Islamic countries. Law enforcement agencies and relevant institutions are interested to filter/ detect hateful speech from social media such as Facebook and Twitter; news websites and blogs/ discussion forums. Recently, in Pakistan, former Governor of Punjab was murdered by his own security guard~\footnote{https://en.wikipedia.org/wiki/Salmaan\_Taseer}. He shared his views on Pakistan Blasphemy Law. His statement was one of the most discussed statements on Facebook and Twitter and other media sites, on those days which ultimately provoked one of his security guard, Mumtaz Qadrito shoot him dead on 4-January 2011. After his demise, policy makers are very much interested to identify those tweets or messages from social media which may incite violence or hurt individual or group sentiments. Since, the data of tweets or messages is unstructured and most of the time multilingual so researchers are very must interested to come-up with some robust framework or set of robust features from these data which may be used for classification (Hate or not-hate speech) or mining of patterns which might be helpful to decision makers. 

Freedom of speech is everyone’s right but some people misuse their right to promote hate and incite violence~\cite{Hate1}. Social websites such as Facebook and Twitter have strict policies to avoid hate speech. They provide feedback or report options on every tweet or message which of course requires user (manual) interaction. If a group of people agree with the hate contents, then the reporting ratio of that tweet or message as hate would be very less. Therefore, it is important to identify hateful speech automatically rather than someone reports it as hateful. Recently, mining or detection of hateful speech is trending among researchers. Some sparse features or tuned parametric classifiers are proposed. However, all reported work uses some existing techniques of classification. All these techniques heavily depend on state-of-the art features. Anyone can easily deceive these frameworks. For example, instead of using slang words in English, a user may write that word in Roman of his/her native language. Then at the feature extraction phase, features are deceived. Therefore, understanding the context of discussion and use of multilingual features are necessary. The aim of this research study is to identify more negative classes from social media comments and tweets using machine learning. The interesting patterns which can be helpful for decision making. We have exploited un-structured behavior of speech delivery by identifying negative comments using sentiment analysis, later these negative comments are further associated to 8 new classes. 

\section{Literature Review}
The revolution of social media has made possible to bring the opinions of people around the globe. Some opinions prevail positivity and some negativity. It is quite unfortunate that negative opionions get viral earlier which compelled the researchers to devise a way restrict this overwhelmingly increasing negativity on the web.
In a research study the tweets were analyzed for the purpose of classification of data and sentiments from Twitter. Extraction of information from tweets were obtained by knowledge base knowledge extraction. Furthermore, the obtained knowledge was strengthened by domain distinct seed-based embellishment facilities. The suggested procedure promotes the extrication through keywords, items, metonym, and native speech form comments used for sentiment analysis and regulation. They tested their proposed methodology on selection of 40,000 tweets. The recommended technique has worked superior than the current system in terms of sentiment analysis and classification \cite{batool2013precise}. Another study represents the sentiment classification system that refers to SemEval (semantic-evaluation) 2015 task 10. The subtask according to the message-level classification method, a maximum prominent F1-scores on 3 sets out of 6 testing sets was assessed. To forecast the sentiment label for tweets, a two-stage classifier was designed. The advancement in their study is enhancement of some deep learning techniques to undoubtedly extract knowledge for sentiment based on message level tasks on word clusters, n-grams, lexicons and twitter data. Besides, nominating the polarity support trick which raises the achievements of system \cite{dong2015splusplus}. An empirical study outlines the experiments on measuring the affection of emotion from tweets using a GRS. This structure incorporates lingual, semantic word impacting aspects, exercising on general regression and definitely merging the perfect observing structure to design all at once \cite{duppada2017seernet}. Opinion Mining (OM) is a current alternate development at the crisis of information retrieval and computational syntax which is bothered not with the case a document is about, but with the opinion it is declared. Various types of application are using the technique of OM, reviews about the products, comments about political candidates convey in online gathering, pasturing from exposing users, to client affiliation administration. The current research has approved to automatically resolve the “PNpolarity” of subjective terms for the evocation of opinions from text, i.e. analyze if a phrase that is pointed of opinionated complacent has a positive or negative significance. Study on resolving if a phrase is surely a point out of opinionated content (intuitive phrase) or not (detached phrase) has been, rather enough extra sparse. SENTIWORDNET, a literal ability in which each WORDNET synonym set is correlated to three numeral result Obj(s), Pos(s) and Neg(s), defining how positive, negative and objective a phrase consists of in the synonym sets was proposed. The technique used for SentiWordNet development is established on the quantitative analysis of the glazes combined to synonym sets. For the delegation of semi-supervised synset classification vectorial terms used for resulting. SentiWordNet in open source and applicable for the reason of exploration and provides a web-based user interface. It was considered that SentiWordNet has appropriate tool for the application such as opinion mining \cite{esuli2006sentiwordnet}. A connected path with Sentiment Analysis and Social Network Analysis was studied. They have tested to accomplice a sentiment to the nodes of the graph displaying the social networks. They illuminate the accession to the problem, with both the system architecture, and then examine first results. On the other side, the network geology can investigate and then uncover some erroneous results of the Sentiment Analysis \cite{fornacciari2015social}. The new emergence of the technique of emotion to classify Twitter tweets accordingly has been deliberated in a research study. With respect to the question word, the messages are scored as positive or negative. Before purchasing any product the public sentiment about that product will be useful for the consumers by first determining the sentiment of that product. On microblogging services like Twitter, there is no past research on classifying the sentiment of messages. Using distant supervision for sentiment of twitter messages for classifying the Machine Learning Algorithms (MLA) were utilized. The training data contains twitter messages with emoticons, is used as noisy labels. This kind of data is generously available and achieved over automated means. When trained with emoticon data it shows that MLA (such as Maximum Entropy, Naïve Bayes, and SVM) veracity of more than 80 words of the sentence and their correlate attitude are known as sentiment thesaurus able for use in sentiment analysis. The research maps for the completion of a replicated digital quantity reflect the spectrum of affection declared in a statement. Related through grouping function is finding through emotions expressed on a comment, the current work can bring extra delightful (re-volute) sentiment analysis. A framework that adopts a model of LSTM (Long Short Term Memory) direction and a convolutionary neural network structure to integrate the tasks of counteraction was proposed. Linking these two models, prevision action acknowledges the overall report in a tweet and limited relevant information. The suggested practice graded sixth amid  twenty-one teams in charge of person Correlation Coefficient. For representing the emotional states two major approaches categorical and dimensional representation were used. Certain distinct collections such as positivity and negativity represent the emotional states using the categorical approach. In various sentiment applications emotions are described as main emotions (anger, fear, sadness) these are successfully adopted to sentiment applications. For more fine-grained (real-valued) results for better sentiment analysis, an approach known as the dimensional approach can be used.  It would be able to have the scale of the strength of emotions derived from the texts by more practical (well-informed) sentiment use (tweets). The study describes a technique known as the deep learning approach for determining the scale of emotions from tweets. The suggested model associates the LSTM, CNN for concealing the general instruction snare by least short-term memory and highest capable components of CNN \cite{he2017yzu}. A system based on Maximum Entropy Classifier for sentiment polarity and the detection of aspect category was proposed. For opinion target extraction the system uses (CRF). Under 9 experiments the updated results are completed and in two attempts with unconstrained structures. The work is divided into three basic three work (SB1) decision level, (SB2) verse level, and (SB3) ABSA outfitting. Advanced Application supported this project L01506. Resources for computational work were procure by the “Projects of Large Research, Development \cite{hercig2016uwb}. An other attempt with SemEval, commenced with donation in SemEval2016 task7. A web-based searching methods for English  and Arabic opinion intensity indicators for defining sentiment intensity for English and Arabic wordings. Their work was placed, firstly on a crowd of classic sentiment lexicons using SentiWordNet on 140 lexicons.  Secondly, the capacity to understand the coherence of sentences with pre-ordained negative and positive terms on online search engines was enabled. By using web search engines (Google Search API) elevates the achievements on idioms created from adverse polarity terms. Implementing our model to the three sub-tasks, experience to other challengers, model established on the supervised entrance with much higher costs than others. Despite the results were supportive, more exploration was prescribed in both languages, regarding the selection of positive and negative words by once correlated to idioms, they prepare it extra negative or extra positive \cite{htait2016lsis}. The task contribution in SemEval 2017 task 4, in this study, defines the “Sentiment Analysis in Twitter” also includes the subtask based on MPC (Message Polarity Classification) for two languages English and Arabic. A series of sentiment seed word was prepared for tweets according to study design. Cosine similarity is measured among the word influencing productions for the sentiment association for seed word and other products (word2vec). Dataset of tweets was accessible online and seed words were extracted from the dataset. The analysis on these seed word shows the serious enhancement in results on polarity classification of tweet messages. Also exploring the polarity classification on different seed words like Turney and Littman’s shows much fewer results compared to their study design \cite{htait2017lsis}. The model used in Tweets to extract the effect was discussed in a research study. Their proposed model joins two contrasting techniques. The first model was termed as N-Stream Convolution Networks that is a deep learning mechanism, whereas the second framework was XGboostregressor placed on to set of impacting and thesaurus based features. The model was graded according to the tasks with the help of the ensemble technique, this model performed accurately. The result of N-Channels ConvNet was much similar to the ensemble model \cite{jabreel2018eitaka}. The strategy was introduced to deal with sensitive tweets. The methodology was aimed at forming lingual characteristics, achieving ideal consolidation features to render a tweet's ultimate vector reappearance, and training postulate regression, describing the sensitivity scores to match the tweets' vector GBR and neural network regression layout. System recommended individual models for each emotion and obtained average person and spearman correlation \cite{john2017uwat}. Companies are expanding the acceptance of social media technologies after all the innovations in social media, using Twitter to scan for buyers or YouTube to view product categories. Most of the business firms today prefer social media as an online marketing tool to agrandize their sale. A research study, based on client sentiment density in social media was conducted. The study covers system architecture for both systems resolving the best attributes that the system should have in structure to build a data analysis application. The researcher were suggested a new architecture which placed on the best features from both modern appointed systems \cite{khairuddinmodel}.  The analysis were done for addressing the strength of a single text or multi-text emotion. Three fields were collected from single text and multi-text: generic English, English Twitter, and Twitter Arabic. It was specified by the best-worst scaling method for expounding. The idioms often consist of adverbs, solutions, modules, and distance, and more idioms are composed of contrasting polarities of text. From all three domains collected for the dataset that comprise multiword idioms and their essential words, these two were basically explained for actual admired sentiment anxiety grades \cite{kiritchenko2016semeval}. Collection of tweets from Twitter and discovering the emotion intensity (or degree) of tweets is still a work in growing. The tweets can be kind of an event, a political party, a political person the degree (intensity) of emotions from tweets can be a useful statement or data for predicting the future on behalf of today’s statement. In NLP application the task of identifying the emotions recognizing the emotion intensities from the text is the area that is still growing \cite{koper2017ims}. In an article, the groundbreaking model was explained based on an architecture that computes words' continuous vector representation against large data sets. The nature of these delegation was systematic by a text similarity checking task, and the conclusion were related to the already best achieving frameworks placed on various sort of neural networks. An immense advancement inefficiency at a lesser computational cost was discovered. It takes less than a day to apply the high-quality word-to-speech vectors against 1.6-billion-word data set. When checking syntactic and semantic word similarities in their test it was revealed that these vectors prepare state-of-the-art completion. Working with DistBelief distributed framework, it should be feasible to train the CBOW (continuous bag of words) and Skip-gram models alike on corpora with one trillion words, for the mostly limitless size of the vocabulary \cite{mikolov2013efficient}. The tasks of discovering the emotions from text through tweets can determine, how the person is feeling and the degree of their emotions whether the person is sad or too sad. The level of emotion and the challenge shared by several communities collaborating on the tasks that remove the emotion from the text serve to advance our understanding of how we bear extra or lower emotions within various languages and how it can even inform us the degree of our emotions with the help of NLP applications. We use different languages for communication not only for understanding the emotions but also the intensity of that emotions that we are feeling, intensity is the degree of emotion that a person is feeling such as the person is angry, the person is sad, the person is in depression \cite{rani2020comparative}. The study focuses on identifying the degree of emotions sense through the use of comments. The first collection of tweets was created using the best-worst scaling technique (BWS) to explain the fundamental emotional intensities of three emotional anger, sadness, fear. The interpretation suggested a major concrete severity.  Knowledge was separated into three phases train the model, development, and testing for competition. The competing team used knowledge-based techniques, tools, structure, basics that are appropriate for the task. The shared task and the emotion intensity dataset are serving to advance our understanding the transferring less and deeper affection through languages. Using the terminology only to feel the emotions we are feeling is not sufficient but also the degree of that emotions. Including nonstandard language such as emoticons, emoji’s, gravely spelled words, and  hashtagged (happy) words, Twitter consists a huge end- user sordid which encompasses prosperous dextral vocabulary. Through tweets, a user can schlepp his/her emotions, opinion and attitude. The study delibrated upon detecting the degree of emotions sense by announcer of a tweet. Using BWS and crowdsourcing the researcher design the emotion intensity datasets. To show that affect of lexicon a standard regression structure were built and conducted an analysis \cite{mohammad2017wassa}.

\section{Methodology}
We have collected the tweets and targeted only Pakistan region. Since, the national language of Pakistan is Urdu, so the tweets from Pakistan are mostly in three formats: English, Urdu fonts, and roman Urdu. Using WordNet as dictionary, only those tweets are extracted that are in English. 

As explained above, we extracted the tweets and filtered only the tweets in English. We used state-of-the-art sentiment analysis using Flair API~\footnote{https://github.com/flairNLP/flair}. We extracted 10000 tweets and then classified the tweets into positive or negative classes~\footnote{https://rileymjones.medium.com/sentiment-anaylsis-with-the-flair-nlp-library-cfe830bfd0f4}. Out of 10000 classes, only 2577 were negative classes.

Later, we requested three experts to manually annotate these 2577 tweets into 8-classes. The final label of the class is chosen if two of the experts have same annotation. The distribution after annotation is shown in the  Table~\ref{tab:1}. The corruption class is moderately imbalanced. The dataset is then divided into two sets, training and test test. The ratio of test set is set to 33\%.

\par Since, there are more than 2 classes which makes the problem multi-class classification. Total four classification algorithms are used on the dataset, the algorithms include SVM (linear support vector machines), Na\-ıve Bayes, BOW (bag of Word), and logistic regression. All these algorithms are implemented using Python APIs such as NLTK which is widely used for  natural language processing.
Susan Li tutorial is followed for the implementation of all models for the experiments~\footnote{https://towardsdatascience.com/multi-class-text-classification-model-comparison-and-selection-5eb066197568}. We have placed our dataset and scripts online for the reimplemntation and use which can be requested from the principal author. 

\section{Results and Discussion}

\begin{table}
	\caption{Dataset distribution}
	\centering
	\begin{tabular}{lllll}
		\toprule
		\multicolumn{4}{c}{Part}                   \\
		\cmidrule(r){1-4}
		Classes & Training set & Testing set &  Class distribution\\ \\
		\midrule
		Politics       & 395 & 131 & 20.41 \\
Injustice      & 297 & 99  & 15.37 \\
Crime          & 252 & 84  & 13.04 \\
Economic       & 234 & 76  & 12.03 \\
Failure        & 209 & 70  & 10.83 \\
Terrorism      & 202 & 68  & 10.48 \\
Social Aspects & 195 & 65  & 10.09 \\
Corruption     & 150 & 50  & 7.76 \\
		\bottomrule
	\end{tabular}
	\label{tab:1}
\end{table}

The classification accuracy of different models used for experiments are shown in Table~\ref{tab:2}. It can be seen that logistic regression achieves marginally better performance compared to the others. Whereas, SVM is very close. Using quadratic or cubic kernel of SVM can improve the accuracy. Since, few of the classes were slightly imbalanced. Therefore, precision and recall measures are also used for further investigation of each class. 

\begin{table}
	\caption{Accuracy of different Models}
	\centering
	\begin{tabular}{lllll}
		\toprule
		\multicolumn{2}{c}{Part}                   \\
		\cmidrule(r){1-2}
		Models & Test Accuracy \\ \\
		\midrule

Linear SVM & 0.7093\% \\
Logistic Regression & 0.7144\% \\
Naive Bayes & 0.6136\% \\
BOW with Keras & 0.5374\%  \\

		\bottomrule
	\end{tabular}
	\label{tab:2}
\end{table}

The	results	of precision, recall and F1 measures on Na\-ıve Bayes, Logistic Regression, Linear SVM (support vector machines), BOW (bag of words) with Keras, are shown in Tables~\ref{tab:3}, \ref{tab:4}, and \ref{tab:5}. 

\begin{table}
	\caption{Logistic Regression accuracy}
	\centering
	\begin{tabular}{lllll}
		\toprule
		\multicolumn{5}{c}{Part}                   \\
		\cmidrule(r){1-5}
		Classes & Precision & Recall & FI-score & Support \\ \\
		\midrule
		Politics       & 395 & 131 & 20.41 \\
Terrorism	& 0.59 & 0.38	& 0.46	& 63 \\
Corruption	& 0.71 & 0.71	& 0.72	& 100 \\
Failure	& 0.77 & 0.74	& 0.76	& 105 \\
Politics	& 0.66 & 0.61	& 0.63	& 76 \\
Injustice	& 0.79 & 0.89	& 0.84	& 125 \\
Social Aspects	& 0.70 & 0.79	& 0.74	& 160 \\
Crime 		& 0.65 & 0.61	& 0.63	& 69  \\
Economics	& 0.71 & 0.71	& 0.71	& 76 \\
Avg/Total	& 0.71 & 0.71	& 0.71	&774 \\
		\bottomrule
	\end{tabular}
	\label{tab:3}
\end{table}

Experiments show that the logistic regression has better performance than linear VSM, N\-ıve Bayes, and BOW with Keras. Table~\ref{tab:3} to ~\ref{tab:5} depict the results of all classes with test accuracy. 

\begin{table}
	\caption{Linear SVM}
	\centering
	\begin{tabular}{lllll}
		\toprule
		\multicolumn{5}{c}{Part}                   \\
		\cmidrule(r){1-5}
		Classes & Precision & Recall & FI-score & Support \\ \\
		\midrule

Terrorism	& 0.79 & 0.35	& 0.48	& 63 \\
Corruption	& 0.68 & 0.73	& 0.70	& 100 \\
Failure	& 0.74 & 0.70	& 0.72	& 105 \\
Politics	& 0.63 & 0.59	& 0.61	& 76 \\
Injustice	& 0.76 & 0.90	& 0.82	& 125 \\
Social Aspects	& 0.70 & 0.82	& 0.76	& 160 \\
Crime 		& 0.75 & 0.57	& 0.64	& 69  \\
Economics	& 0.67 & 0.70	& 0.68	& 76 \\
Avg/Total	& 0.71 & 0.71	& 0.70	& 774 \\
		\bottomrule
	\end{tabular}
	\label{tab:4}
\end{table}

\begin{figure}
	\centering
 \includegraphics[width=0.75\textwidth]{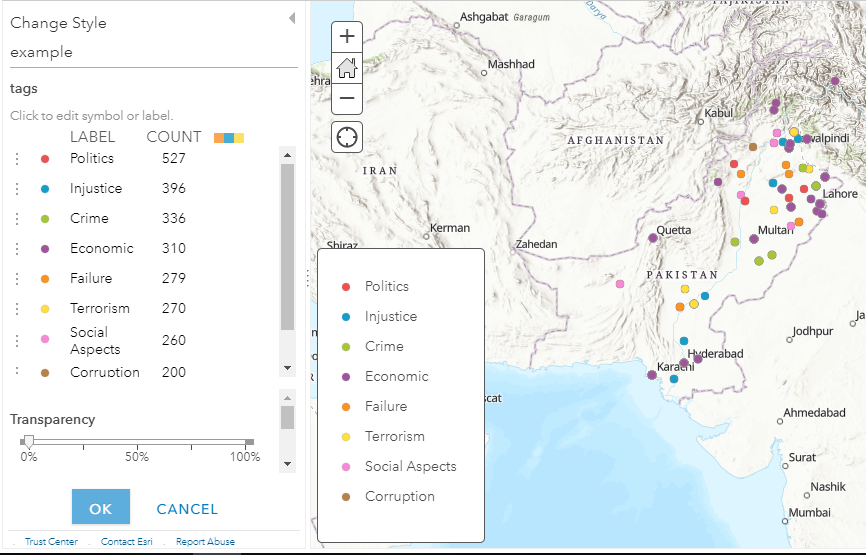}
	
	\caption{Geo mapping of negative classes based on the frequencies.}
	\label{fig:map}
\end{figure}

\begin{table}
	\caption{Naïve Bayes}
	\centering
	\begin{tabular}{lllll}
		\toprule
		\multicolumn{5}{c}{Part}                   \\
		\cmidrule(r){1-5}
		Classes & Precision & Recall & FI-score & Support \\ \\
		\midrule
Terrorism	& 1.00 & 0.03	& 0.06	& 63 \\
Corruption	& 0.76 & 0.64	& 0.70	& 100 \\
Failure	& 0.94 & 0.57	& 0.71	& 105 \\
Politics	& 0.92 & 0.32	& 0.47	& 76 \\
Injustice	& 0.79 & 0.86	& 0.82	& 125 \\
Social Aspects	& 0.40 & 0.98	& 0.57	& 160 \\
Crime 		& 0.69 & 0.32	& 0.48	& 69  \\
Economics	& 0.87 & 0.51	& 0.64	& 76 \\
Avg/Total	& 0.78 & 0.61	& 0.59	&774 \\

		\bottomrule
	\end{tabular}
	\label{tab:5}
\end{table}

we also observed that as the geographic location changes, the issues of the people also changed. After learning the models, we visualized the classes frequencies with the geo location of the people who tweeted the negative tweets. The  ArcGIS tool is used to map the location of the people based on the frequencies of the classes distribution. The Figure~\ref{fig:map} shows locations of all classes as in a map. The cities where education ratio is higher are having negative feelings on Economic matters, whereas, the remote areas of Balochistan, which is lowest ranked in education, are facing social aspect based negativity. It can also be seen in the Figure that, political grievances are among the people who live in country capital and nearby cities. 

\section{Conclusion}
The use of social media to share your feelings are now very common. People share their happiness, sadness, achievements, or grievances on their social media accounts or groups. The different trends on social media can play very positive role or even sometimes it can go very opposite. In this research study, we use sentiment analysis to classify tweet into positive and negative classes. Later, we further associate negative class to sub-negative classes. Based on the experts opinion, the negative class is further divided into 8 more classes that include \textit{terrorism, politics, corruption, injustice, failure, crime, social aspects, and economic}. New dataset is created and different machine learning models that include linear SVM (support vector machine), naive Bayes, logistic regression, BOW (bag of words with Keras), are evaluated. 
At the end, based on these negative mapping with the classes, the geo location of Pakistan are highlighted where one can observe which area of Pakistan is facing what kind of issues. In future work, we are extending our framework for Urdu fonts and roman Urdu so that we automatically analyze the people opinion related to negativity in large scale.

\bibliographystyle{unsrtnat}
\bibliography{references}  






\end{document}